# Offline Handwriting Signature Verification: A Transfer Learning and Feature Selection Approach

Fatih Özyurt[1*], Jafar Majidpour[2,3], Tarik A. Rashid[4], Canan Koç[1]

[1] Department of Software Engineering, Faculty of Engineering, Firat University, Elazig, 23119, Turkey
[2] Department of Software and Informatics Engineering, University of Raparin, Rania, Kurdistan Region, Iraq
[3] Department of Computer Science, Faculty of Science, Soran University, Soran, Kurdistan Region, Iraq
[4] Computer Science and Engineering Department, University of Kurdistan Hewlêr, Erbil, Iraq

Corresponding Author Email: fatihozyurt@firat.edu.tr

## ABSTRACT

Handwritten signature verification poses a formidable challenge in biometrics and document authenticity. The objective is to ascertain the authenticity of a provided handwritten signature, distinguishing between genuine and forged ones. This issue has many applications in sectors such as finance, legal documentation, and security. Currently, the field of computer vision and machine learning has made significant progress in the domain of handwritten signature verification. The outcomes, however, may be enhanced depending on the acquired findings, the structure of the datasets, and the used models. Four stages make up our suggested strategy. First, we collected a large dataset of 12600 images from 420 distinct individuals, and each individual has 30 signatures of a certain kind (All authors' signatures are genuine). In the subsequent stage, the best features from each image were extracted using a deep learning model named MobileNetV2. During the feature selection step, three selectors—neighborhood component analysis (NCA), Chi2, and mutual_info (MI)—were used to pull out 200, 300, 400, and 500 features, giving a total of 12 feature vectors. Finally, 12 results have been obtained by applying machine learning techniques such as SVM with kernels (rbf, poly, and linear), KNN, DT, Linear Discriminant Analysis, and Naïve Bayes. Without employing feature selection, our suggested offline signature verification achieved a classification accuracy of 91.3%, whereas using the NCA feature selection approach with just 300 features it achieved a classification accuracy of 97.7%. High classification accuracy was achieved using the designed and suggested model, which also has the benefit of being a self-organized framework. Consequently, using the optimum minimally chosen features, the proposed method could identify the best model performance and result validation prediction vectors.

## 1. INTRODUCTION

### 1.1 Background

Today, artificial intelligence has reached a critical point. The methods and data sets used in the studies are the building blocks that make them necessary. Incredibly unique and different data sets make the work more striking. One of these unique and different datasets mentioned is the handwritten signature dataset.

Signatures, such as letters and papers, are included in official documents to affirm identity or agreement. Since ancient times, it has been seen in different forms in various states, professions, and art institutions. Handwritten signatures are usually created with ink or wax. The signature can be a simple "OK," a specific emblem, or even your handwritten name that verifies the document. Handwritten signatures are one of the most widely accepted ways to prove the legal validity of documents.

Handwritten signatures authenticate a person to protect their privacy and security [1]. Verifying whether a signature is real or fake is a critical area of research, and the primary purpose of these systems is to reduce and prevent fraud.

The definition and authenticity of a signature are of critical importance. Signature recognition systems are being developed to authenticate signatures and detect forgeries. These systems are classified as online and offline [2]. The online system uses an electronic tablet and pen attached to a computer to verify dynamic information such as pressure, speed, and typing speed. In offline systems, the signature template comes from an imaging device, and only static data is obtained. Offline signature recognition systems are more valuable and useful. For this reason, the offline signature verification technique was used in the study done in this article.

### 1.2 Motivation

The datasets used in offline signature verification mostly consist of signed signatures based on the signers' names. It is possible that several signers have the same name. Conversely, some datasets exhibit an imbalance, with restricted participants. To address these problems, we compiled a dataset in which all signers signed using their own unique structures. Our dataset is carefully balanced to include an equal number of signatures from each signer.

There is a lot of work in the literature on signatures and verification. In 2014, Kudlacik and colleagues used a dataset of 20 genuine and 20 forged signatures from 40 users. They have achieved an accuracy rate of 99.19% with the fuzzy method applied as a result of the work done [3]. Radhika et al. conducted their work on online and offline signature verification [4]. Their study using the SVM algorithm obtained a 74.04% accuracy rate in offline work. Serdouk et al. proposed a new system called AIRSV for offline signature verification using three datasets: MCYT-75, GPDS-300, and GPDS-4000 [5]. As a result of the work done, they achieved a 76.59% accuracy rate.

In 2023, Longjam et al. proposed a hybrid system called CNN-BiLSTM [1]. GPDS-300, GPDS-Bengali, GPDS-Devanagari, CEDAR, and BHSig260-Bengali, for signature verification in this proposed hybrid system. BHSig260-Hindi and Meitei Mayek datasets were also used. As a result of the study, they achieved 100% accuracy with the CEDAR dataset. 2023, Ren et al. proposed a two-channel and two-flow transformer approach (2C2S) to solve the signature verification problem [6]. At the end of the study, the four datasets used, SUES-SIG, CEDAR, BHSig-B, and BHSig-H, obtained an accuracy rate of 93.25%, 90.68%, 100%, and 72.22%, respectively.

Hafemann et al. proposed formulations for learning characteristics for offline signature verification in their study [7]. Experiments on GPDS-960 have yielded 1.72% ERR, compared to 6.97% in the literature, and this result shows a significant improvement in the latest technology.

Jahandad et al. used the GPDS dataset to achieve 83% and 75% accuracy using the InceptionV1 and InceptionV3 architectures [8]. Tuncer and his colleagues present a new deep signature verification model [9]. The study, which consisted of deep feature creation using transfer learning, iterative minimum redundancy maximum relevance (IMRMR) property selection, and classification phases, used both the CEDAR dataset and the dataset collected by the researchers. At the end of the study, 97.16% accuracy was achieved in the dataset collected and 100% accuracy in the CEDAR dataset. The study by Majidpour et al. proposed a new use of the Generative Adversarial Network (GAN) model as a high-quality data synthesis method to solve the problem of unreadable data in signature verification [10]. In addition to the three architectures used in the study, MobileNet, SqueezeNet, and ShuffleNet, three different high-intensity noises, Salt & Pepper (S&P), Gaussian and Gaussian Blur, were added to the images in the pre-processing phase to make the signature unreadable. The highest success rate was 98.68% using Gaussian Blur and MobileNet.

The study by Banerjee et al. developed an offline signature verification model that does not change the language and is almost equally applicable for both writer-dependent and author-independent scenarios [11]. CEDAR, UTSig, Sigcomp 2011 Dutch, Sigcomp 2011 Chinese, and SigWIcomp 2015 Bengali signature datasets were studied. At the end of all studies, an accuracy rate of 99% was achieved in the Sigcomp 2011 Chinese dataset. An overview of numerous published investigations is shown in Table 1.

**Table 1.** A summary of previous studies

| Year | Researchers | Proposed Method | Data Numbers | Accuracy |
|------|-------------|-----------------|--------------|----------|
| 2014 | Kudlacik et al. [3] | Fuzzy Logic | 1600 | 99.19% |
| 2015 | Radhika et al. [4] | SVM | - | 74.04% |
| 2017 | Serdouk et al. [5] | AIRSV | 1125 genuine, 1125 forged | 76.59% |
| 2017 | Hafemann et al. [7] | SigNet, SigNet-F | GPDS-960 | ERR 6.97% |
| 2019 | Jahandad et al. [8] | InceptionV1, InceptionV3 | GPDS | 83% and 75% |
| 2021 | Tuncer et al. [9] | Transfer Learning, IMRMR | CEDAR, Collected dataset | 100% and 97.16% |
| 2021 | Banerjee et al. [11] | Red Deer Algorithm, proposed meta-heuristic method, Naïve Bayes | CEDAR, UTSig, Sigcomp 2011 Dutch, Sigcomp 2011 Chinese and SigWIcomp 2015 Bengali | Sigcomp 2011 Chinese 99% |
| 2022 | Majidpour et al. [10] | MobileNet, SqueezeNet, and ShuffleNet Salt & Pepper (S&P), Gaussian, and Gaussian Blur | Indic scripts dataset | Gaussian Blur and MobileNet 98.68% |

## 1.3 Contribution

The novelty of this study is that a new signature dataset has been obtained that has not been used before.

The contributions of the proposed approach are:

- 420 distinct signatures, 30 genuine signatures for each class, were gathered for an offline signature collection.

- To improve model performance accuracy and decrease model time complexity, the optimal minimal features for recognizing the signatures should be chosen from a range of 38%, 31%, 23%, and 15% of all features.

- Achieving remarkable results of more than 92% with the fewest features possible via popular machine learning algorithms such as SVM (which includes linear kernels, rbf, and poly), KNN, DT, LDA, and Naïve Bayes.

Identification of the signatures based on different machine learning classifiers with high accuracy.

## 1.4 Organization

Information about the handwritten signature image corpus and the proposed image classification framework is explained in Section 2. Section 3 presents the complete experimental results. Lastly, the conclusion part is given in Section 4.

## 2. MATERIAL AND METHODS

To verify each offline handwriting signature with the highest accuracy, the fewest mistakes, and minimal processing time, an approach combining deep learning, feature selection techniques, and machine learning techniques is presented in Figure 1. Our suggested model's framework consists of the following four steps: (i) dataset and preprocessing include steps like collecting and preparing a private dataset, cropping, and resizing; (ii) feature extraction uses a TL-based MobileNetV2 model; (iii) feature selection uses fine-tuned NCA, Chi_2, and mutual_info; (iv) offline signature verification uses machine learning techniques like SVM with kernel (rbf, poly, and linear), KNN, DT, Linear Discriminant Analysis, and Naïve Bayes. Here are thorough descriptions of each step. The proposed structure is shown in Figure 1.

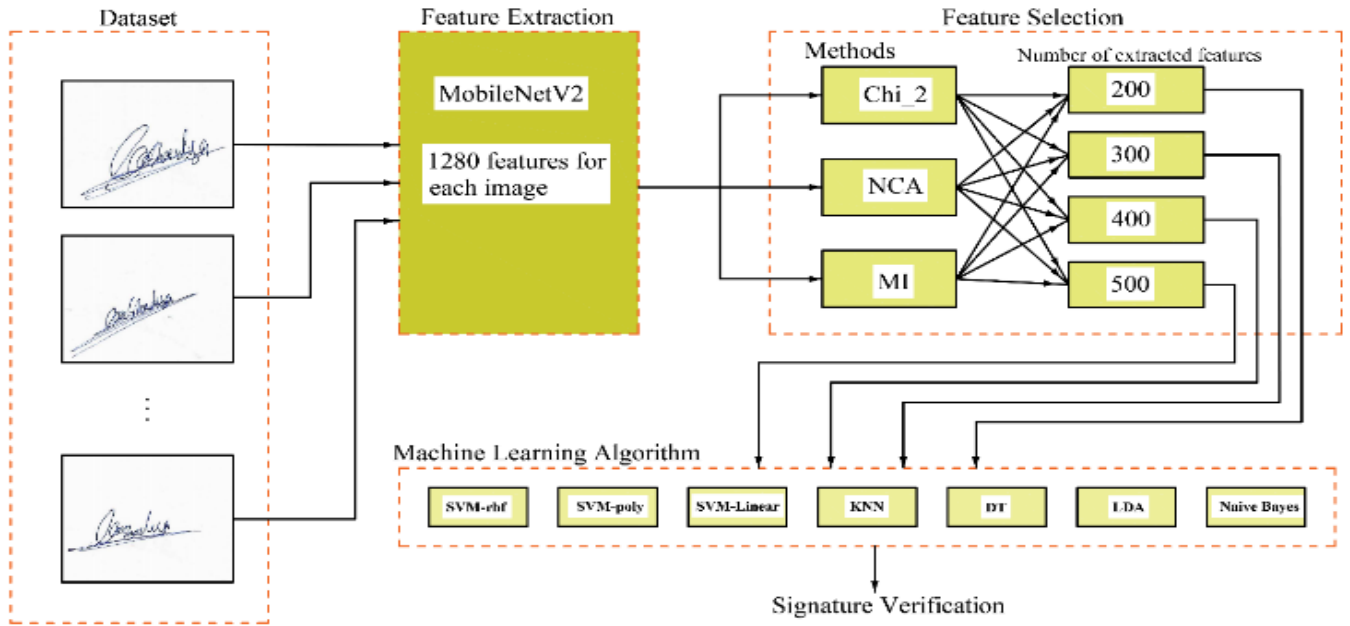

**Figure 1.** The proposed structure

## 2.1 Dataset

12600 images from 420 people were gathered for a private dataset, along with 30 participants' signatures. Students and faculty members of the University of Raparin in Ranya, Iraq, provided all of the signatures. The procedures for gathering our dataset are outlined below:

1. Set up an A4 sheet of paper with a grid that has five rows and three columns so that each individual can sign 15 signatures on each side.
2. Provide each signer with two grid-lined sheets of paper for the required 30 signatures.
3. Each person's signed signatures must all be of the same type.
4. Blue and black pens must be used to sign each signature on the grid paper.

We have a team dedicated to gathering signatures and providing input before, after, and throughout the signing process. Dataset collection took place over the course of two months. A selection of signed signatures is presented in Figure 2.

efficient. In MobileNetV2, the size of the feature maps has been narrowed using 1x1 convolutions. It is one of the facts that the MobileNetV2 architecture is known to have fewer parameters in its layers than other architectures [15].

The main advantage of the MobileNet architecture is that it performs relatively few computations compared to the traditional CNN model. With two different parameters used in the architecture, it is possible to control the details and delay time of the objects in the image. If the values entered in these parameters are kept low, they can have the same effect as the minimum number of properties, such as Palmprint recognition. As a result, it is possible to simplify or make the architecture complex [16]. This is one of the main advantages of the MobileNetV2 architecture [17].

The MobileNetV2 architecture is one of the architectures where the deeply detachable convection method is preferred. If this architecture is compared with the first version, it is seen that the number of weights is reduced further by narrowing the output channels. In addition, a layer of point convection has been imported before the deeply detachable convection, further improving its performance.

### 2.3 Feature Selection

Feature reduction and feature vector expansion are the goals of this subsection. We employ the three most well-known feature selectors. NCA [18], Chi2 [19], and MI [20] are the feature selectors in question. The following sections expand on the function of each feature selector.

#### 2.3.1 NCA

For choosing discriminative features and for maximum classification accuracy, NCA is a feature selection method based on distance. Equation (1) provides the normalized equation for computing the NCA feature selection for a multi-class SEC problem.

$$S = \{(x_i, y_i), i = 1, 2, \dots, n\} \qquad (1)$$

where, $c$ is the number of emotional class labels, $x_i$ signifies the feature vectors, $y_i$ denotes the class labels of emotional features, and $n$ is the number of observations [21]. To put it simply, the objective is to give the classifier $f : R^p \to \{1, 2, \dots, c\}$ relevant features (from the feature extraction segment of the model) so that it can perform a classification $f(x)$ for the ground truth (label) $y$ of $x$.

#### 2.3.2 Chi-Square algoritm

It is based on whether the difference between observed and expected frequencies makes sense. It is used in the qualitatively specified analysis [22]. Which is used for the chi-square test,

- In the test of whether there is a difference between two or more groups,
- In the test of whether there is a link between the two variables,
- In the inter-group homogeneity test,

Test whether the distribution obtained from the sample conforms to any desired theoretical distribution (compatibility goodness test).

#### 2.3.3 Mutual Information (MI)

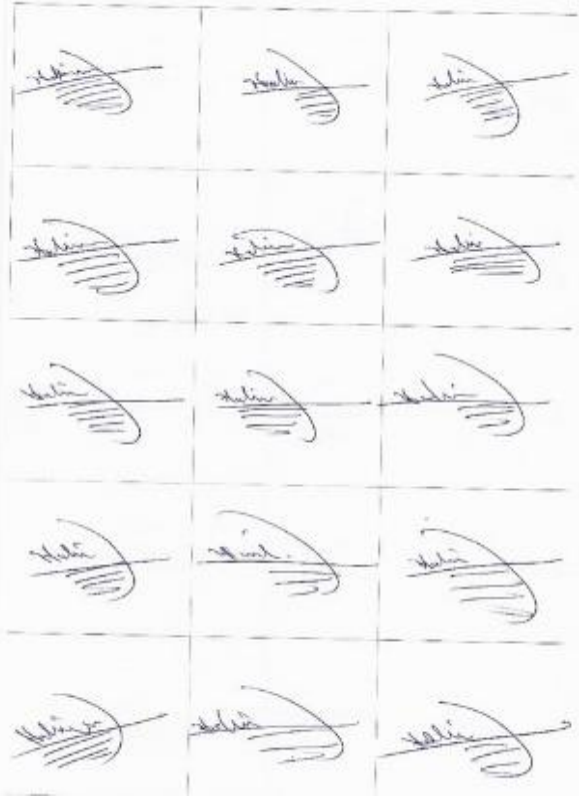

**Figure 2.** A selection of signed signatures

### 2.2 Transfer learning model

A large dataset is necessary for training CNNs to obtain the necessary accuracy, but there are some situations when the challenges of putting up a large dataset may compromise the model's performance accuracy. Real-world training and testing data pairs are notoriously challenging to obtain [12]. "Transfer learning" was proposed as an answer to this problem. To put it another way, transfer learning is a method of machine learning where a trained model is used as the foundation for a model applied to a different problem. It is possible to model the second task quickly by using an optimized model for the first task and then applied to the second. The outcomes of applying transfer learning to a new activity are significantly more spectacular than training with little to no data [13]. Our suggested approach uses the MobileNetV2 transfer learning model for feature extraction, which is described in the section below.

#### 2.2.1 MobileNetV2 as feature extraction

MobileNet is a Convolutional Neural Network designed for mobile and embedded vision applications. The MobileNet architecture aims to develop deep learning applications on mobile and embedded systems with lower data processing capabilities than computers [14]. Although designed with mobile environments in mind, it is also a preferred deep learning architecture in desktop environments. This architecture has been driven by researchers' drive to create small but efficient deep-learning architectures with fewer training parameters. In extracting features from images with convolutional filters, it uses the technique of Depthwise separable Convolutions instead of the standard convective operation. In 2019, MobileNet's architecture was developed and became MobileNetv2, which will be faster and more In feature selection, MI is a statistical technique. It measures the degree to which two variables (a and b) are interdependent. Through the use of the other random variable, it assesses the "measure of data" amassed regarding one arbitrary variable [23]. The MI between two discrete random variables, a and b, is calculated using Equation 2.

$$I(A,b) = \sum_{b \in B} \sum_{a \in A} p(a,b) \log \left( \frac{p(a,b)}{p(a)p(b)} \right) \qquad (2)$$

where, $p(a,b)$ represents the joint probability function of $A$ and $B$, and $p(a), p(b)$, respectively, represents the marginal probability distribution functions of $A$ and $B$.

## 2.4 Classification

Seven well-known classifiers are used to verify the collected offline handwritten signature: SVM [24] with the kernel (Radial Basis Function, polynomial, Linear), KNN [25], DT [26], LDA [27], and Naïve Bayes [28]. The specifics of each classifier are detailed in the subsections below.

### 2.4.1 Support Vector Machine (SVM)

Support Vector machines are a supervised learning method often used in classification problems. Support Vector Machines (SVMs) are the decomposition and classification of points on the plane by a straight or hyperplane [24]. It is suitable for small or medium-sized datasets and needs to be scaled.

The purpose of classification problems is to decide which class the future data will be in. To do this classification, a line separates the two classes, and the region between ± 1 of this line is called the margin. The wider the margin, the better the separation of two or more classes.

RBF, which stands for Radial Basis Function Kernel, is an extremely powerful kernel that is utilized in SVM. RBF is more sophisticated and efficient than linear or polynomial kernels because it may combine multiple polynomial kernels multiple times of varying degrees to project non-linearly separable data into higher dimensional space and separable using a hyperplane.

An SVM kernel, known as a polynomial kernel, maps the data into a higher-dimensional space using a polynomial function. This is accomplished by taking the dot product of the polynomial function in the new space and the original space's data points.

The linear kernel—also known as the "Non-kernel"—is the most straightforward of all the kernels. The inner product of x and y with an optional constant term c is all that exists when this kernel is technically utilized because the data isn't projected onto higher dimensions.

### 2.4.2 K-Nearest Neighbours (KNN)

The nearest neighbor algorithm is expressed as a supervised machine learning method in which the class (learning cluster) and the nearest neighbor (element) are to be classified according to the k value (similarity) [25].

General classification algorithms (models) use this classifier on every data value within the system by creating a classifier within their solutions. Compared to these classification algorithms, the KNN algorithm classifies the values for each value by creating a classifier over a set of neighbors closest to the corresponding value.

The K-nearest neighbors algorithm (KNN) predicts two fundamental values [29].

- Distance: The distance of the point to be estimated from other points is calculated. For this, the Minkowski distance calculation function is used.
- K (number of neighbors): The number of neighbors is calculated from the nearest number. The value of K directly affects the result. If K is 1, the probability of overfitting is very high. If it is too large, it gives very general results. Therefore, estimating the optimal K value is the main issue in the problem.

Three indicators are commonly used to measure the performance of a model produced with the KNN (K-nearest neighbors) algorithm.

- Jaccard Index: The ratio of the intersection set of the correct prediction set and the actual value set to their junction set, taking values between 1 and 0. 1 is the best.
- F1-Score: Calculated from precision and recall values calculated from the confusion Matrix. F1-Score = 2*((Pre*Rec)/(Pre+Rec)))) is a value between 1 and 0. 1 is the best.
- LogLoss: At the end of the logistic regression, the LogLoss value is calculated from the probabilities of the predictions. It is valued between 1 and 0. Unlike the above two values, 0 means the best performance.

### 2.4.3 Decision Tree (DT)

A DT is a Supervised Machine Learning Algorithm that uses a set of rules to make decisions like humans do [26]. Classification trees are tree models in which the goal variable can take a discrete set of values; in these tree structures, leaves indicate class labels, and branches represent feature conjunctions that lead to those class labels.

### 2.4.4 Linear Discriminant Analysis (LDA)

LDA is a type of linear combination. This mathematical operation uses a variety of data elements and applies functions to this set to analyze multiple object or item classes separately [27].

Fisher's linear discriminant, flowing from linear discriminant analysis, can be helpful in areas such as image recognition and predictive analytics in marketing. LDA is based on looking for a linear combination of variables that best differentiate between good class (goals). Fisher describes the score function. According to the score function, the problem predicts linear coefficients that maximize the score [30]. Calculating the Mahalanobis distance between the two groups is the best way to determine discrimination. The fact that the Mahalanobis distance is smaller than three means that the probability of misclassification is small.

### 2.4.5 Naïve Bayes

Bayes' theorem is a crucial subject in probability theory [28]. This theorem shows the relationship between conditional and marginal probabilities within the probability distribution for a random variable.

The NB classifier is based on Bayes' theorem. It is a lazy learning algorithm that works on unstable data sets. The algorithm calculates the probability of each case for an element and classifies it according to the one with the highest probability value [31]. They can do very well with a bit of training. If a value in a test set has a value that cannot be observed in the training set, it gives a probability value of 0, which means it cannot make a prediction. This is often referred to as zero frequency. Corrective techniques are used to resolve this situation. One of the most straightforward correction techniques is known as Laplace prediction. Examples of uses include real-time forecasting, multi-class forecasting, text classification, spam filtering, sensitivity analysis, and recommendation systems.

### 2.4.6 Evaluation Criteria

Each classification's effectiveness is evaluated using a variety of observational error measurements. One indicator of its efficacy is how accurately a classification model can categorize data. The frequency with which our model delivers accurate predictions is referred to as "accuracy" in this context. Technically speaking, being precise means:

$$Accuracy = \frac{TP+TN}{TN+TP+FN+FP} \qquad (3)$$

where TP= True Positives, TN = True Negatives, FP = False Positives, and FN = False Negatives.

The percentage of pertinent examples among all the ones found is called precision, which is often referred to as a positive predictive value. It gauges the proportion of projected positive classes that contain members of that class. The equation (4) that follows defines precision.

$$Precision = \frac{TP}{TP+FP} \qquad (4)$$

Recall is occasionally used to demonstrate or evaluate how well a test can conclusively exclude the presence of an illness or disease state. This test accuracy indicator is essential in situations where a false positive could result in significant financial loss. Recall is defined by equation (5).

$$Recall = \frac{TN}{TN+FP} \qquad (5)$$

The F-score, also known as the F1-score, measures how well a model fits the provided data. The F-score, which is calculated, is the harmonic mean of the model's recall and accuracy. To calculate an F1 score, use Equation (6):

$$F1 - Score = \frac{2TP}{2TP+FP+FN} \qquad (6)$$

## 3. EXPERIMENTAL RESULTS

This work aims to use deep learning and machine learning methods to verify offline handwritten signatures. There are three steps in the primary method: In the beginning, we acquired a large data set of 420 distinct individuals, each of whom had 30 identical signatures taken. The second step is to decide which features, based on the fundamental composition and morphological texture of the signature images, will be most helpful, and the third step is to reduce error and boost the accuracy of signature verification.

Each signature image, as is well known, contains few details and is only one color, making it challenging to verify and adding to the complexity. Several preprocessing, feature extraction, feature selection, and classifier methods have been combined for this study, and they are evaluated using a variety of criteria. The preprocessing, training, and testing specifics come before the system's actual implementation in this section. The experiments that employ the suggested strategies are then displayed and reviewed.

### 3.1 Preprocessing

The total number of images includes 12,600 images taken by 420 people, each scanned and stored separately. That each individual is a distinct class, each A4 page had 15 signatures; therefore, we automatically cropped and saved them using MATLAB's Crop function. Each signature was then saved as a distinct image and saved in the class associated with the relevant person. In the final phase, we resized the images to 500 x 600. Figure 3 shows a cropped signature ready to be saved as one image.

We personally reviewed and cropped the images to look for these occurrences because, during automatic cropping, it's possible that certain signatures were removed from the boundary and other signature parts were destroyed.

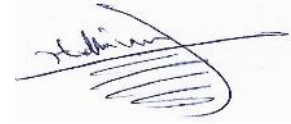

**Figure 3.** A cropped signature and save it as one image

Every stored image is split into testing and training sets. 20% is used to assess the data's performance, while the remaining 80% is used to train the models. The cross-validation (CV) approach [32] is used to assess the system's correctness. We trained and tested each classifier five times on different sets of data using the fivefold CV approach, dividing the data into 80 percent for training and 20 percent for testing. We averaged each of the five CVs in the batch to get the mean CV.

### 3.2 Training and Testing Phases

Our primary focus is on developing and training a model employing feature selectors, deep learning, and machine learning, with the following goals: Features should be minimized and chosen optimally, and the accuracy of the model's performance should be improved.

Despite these challenges, Tables 2, 3, and 4 demonstrate that a limited number of features can nevertheless achieve a high level of verification accuracy for offline handwritten signature verification. The current proposed approach consists of two steps. In the first step, a MobileNetV2 fine-tuning transfer learning model from the global average pooling layer was used to extract all preprocessed gathered signatures and imagined features. A total of 1280 characteristics were retrieved from each image.

Even if several feature selection methods exist, picking the best one for a certain task is never easy. Three well-known feature selectors, including NCA, Chi2, and MI, were used in our suggested model. Each feature selector received all the extracted features and used them to choose the best and most helpful characteristics. The goals we set for ourselves were 200, 300, 400, and 500 features from each of the feature selectors in the various test models. When each feature group chosen by the feature selectors is sent to machine learning algorithms for signature verification, Tables 2, 3, and 4 show the performance accuracy of each model based on each feature group.

### 3.3 Discussion

In this study, firstly, classification was performed without using any feature selection technique. These classification algorithms were performed using different SVM algorithms, KNN, DT, LDA, and Naïve Bayes algorithms. As can be seen in Table 2, when the accuracy rates are compared, the LDA algorithm has the highest accuracy rate of 92.2%. Likewise, when the precision, recall, and f1 score rates are analyzed, the LDA algorithm has a higher success rate than the other algorithms. Another conclusion drawn from Table 2 is that the accuracy, precision, recall, and f1 scores of the KNN and DT algorithms are equal to each other and the lowest compared to other algorithms. Also, when SVM algorithms are compared among themselves, it is seen that SVM-Linear ratios are higher than the others. SVM-Poly is the SVM algorithm with the lowest rates.

**Table 2.** Experimental results without using feature selection methods

|  | SVM-rbf | SVM-poly | SVM-Linear | KNN | DT | LDA | Naïve Bayes |
|---|---|---|---|---|---|---|---|
| **accuracy** | 0.902 | 0.881 | 0.917 | 0.825 | 0.825 | 0.922 | 0.825 |
| **precision** | 0.913 | 0.897 | 0.93 | 0.849 | 0.849 | 0.939 | 0.825 |
| **recall** | 0.902 | 0.881 | 0.917 | 0.825 | 0.825 | 0.922 | 0.825 |
| **f1-score** | 0.899 | 0.881 | 0.918 | 0.821 | 0.821 | 0.925 | 0.825 |

In the second study, feature extraction was performed using MobileNetV2. For feature selection, the data was classified using Chi-2. It is seen that LDA has the highest accuracy rate when 500 features are initially selected (see table 3). Likewise, when precision, recall, and f1-score are compared, it is seen that LDA is the algorithm with the highest rates. When SVM algorithms are compared among themselves, it is observed that SVM-Linear has the highest rate and SVM-Poly has the lowest rate.

When Table 3 continues to be analyzed, it is seen that LDA is the algorithm with the highest rate when 400 features are selected. Among SVM algorithms, the SVM-Linear algorithm has the highest rate. It is observed that the number of features selected gradually decreases while selecting features. When 300 and 200 features are selected, the LDA algorithm has the highest accuracy, recall, precision, and f1 score ratios. On the other hand, the lowest ratios belong to KNN and DT algorithms with equal ratios.

When Table 3 is analyzed in general, the precision ratio between the selection of 500 features and the selection of 400 features is very close to each other and is above 93%.

In the other study, feature extraction was also performed using MobileNetV2. In this study, NCA was used for feature selection. As can be seen in Table 4, the selected features were again reduced from 500 to 200. The difference in this study compared to Tables 2 and 3 is that SVM-Linear has the highest accuracy rate (94.33%), recall rate (94.33%), precision rate (95.07%), and f1-score rate (94.30%) in the selection of 500 features. In the 500-feature selection study, the SVM-Poly algorithm has the lowest accuracy rate (87.54%), recall rate (90.28%), and f1 score rate (90.16%), while the Naïve Bayes algorithm has the lowest precision.

On the other hand, the SVM-rbf algorithm has the highest accuracy, recall, precision, and f1-score ratios in the studies with 400 and 300 features, respectively. The lowest ratios are obtained in the SVM-Poly algorithm. When the study with 200 feature selections is analyzed, it is seen that the highest rates of accuracy and recall are obtained with the SVM-rbf algorithm. The algorithm with the highest precision and f1 score is the SVM-Linear algorithm.

**Table 3.** Experimental results of the proposed model for multiple models based on the Chi_2 feature selection method on 200, 300, 400, and 500 features

| MobileNetV2 1280 Features | Number of Selected Features | Metrics | Machine Learning Models | | | | | | |
|---|---|---|---|---|---|---|---|---|---|
| | | | SVM-rbf | SVM-poly | SVM-Linear | KNN | DT | LDA | Naïve Bayes |
| Chi_2 | 500 | **accuracy** | 88.89 | 86.47 | 90.12 | 81.31 | 81.31 | 92.50 | 81.31 |
| | | **precision** | 90.31 | 88.58 | 91.73 | 84.18 | 84.18 | 93.74 | 81.31 |
| | | **recall** | 88.89 | 86.47 | 90.12 | 81.31 | 81.31 | 92.50 | 81.31 |
| | | **f1-score** | 88.76 | 86.56 | 90.26 | 80.96 | 80.96 | 92.63 | 81.31 |
| | 400 | **accuracy** | 88.02 | 85.56 | 89.60 | 79.64 | 79.64 | 91.71 | 79.64 |
| | | **precision** | 89.71 | 87.96 | 91.08 | 82.59 | 82.59 | 93.14 | 79.64 |
| | | **recall** | 88.02 | 85.56 | 89.60 | 79.64 | 79.64 | 91.71 | 79.64 |
| | | **f1-score** | 87.97 | 85.65 | 89.66 | 79.12 | 79.12 | 91.81 | 79.64 |
| | 300 | **accuracy** | 86.55 | 83.45 | 87.98 | 78.53 | 78.53 | 89.92 | 78.53 |
| | | **precision** | 88.58 | 86.23 | 90.08 | 82.30 | 82.30 | 91.77 | 78.53 |
| | | **recall** | 86.55 | 83.45 | 87.98 | 78.53 | 78.53 | 89.92 | 78.53 |
| | | **f1-score** | 86.49 | 83.62 | 88.17 | 78.19 | 78.19 | 90.10 | 78.53 |
| | 200 | **accuracy** | 84.25 | 80.67 | 86.23 | 75.12 | 75.12 | 88.53 | 75.12 |
| | | **precision** | 86.56 | 83.93 | 88.58 | 78.92 | 78.92 | 90.41 | 75.12 |
| | | **recall** | 84.25 | 80.67 | 86.23 | 75.12 | 75.12 | 88.53 | 75.12 |
| | | **f1-score** | 84.10 | 80.93 | 86.41 | 74.73 | 74.73 | 88.68 | 75.12 |

**Table 4.** Experimental results of the proposed model for multiple models based on the NCA feature selection method on 200, 300, 400, and 500 features

| | Metrics | Machine Learning Models |
|---|---|---|

| MobileNetV2 1280 Features | Number of Selected Features | | SVM-rbf | SVM-poly | SVM-Linear | KNN | DT | LDA | Naïve Bayes |
|---|---|---|---|---|---|---|---|---|---|
| NCA | 500 | accuracy | 93.93 | 87.54 | 94.33 | 90.28 | 90.28 | 93.10 | 90.28 |
| | | precision | 94.57 | 92.24 | 95.07 | 91.76 | 91.76 | 94.18 | 90.28 |
| | | recall | 93.93 | 87.54 | 94.33 | 90.28 | 90.28 | 93.10 | 90.28 |
| | | f1-score | 93.69 | 88.43 | 94.30 | 90.16 | 90.16 | 93.15 | 90.28 |
| | 400 | accuracy | 97.62 | 64.76 | 97.34 | 95.79 | 95.79 | 96.47 | 95.79 |
| | | precision | 97.47 | 90.15 | 97.43 | 96.55 | 96.55 | 96.89 | 95.79 |
| | | recall | 97.62 | 64.76 | 97.34 | 95.79 | 95.79 | 96.47 | 95.79 |
| | | f1-score | 97.34 | 72.22 | 97.33 | 95.88 | 95.88 | 96.51 | 95.79 |
| | 300 | accuracy | 97.70 | 53.45 | 97.34 | 95.63 | 95.63 | 96.27 | 95.63 |
| | | precision | 97.61 | 76.35 | 97.50 | 96.26 | 96.26 | 96.72 | 95.63 |
| | | recall | 97.70 | 53.45 | 97.34 | 95.63 | 95.63 | 96.27 | 95.63 |
| | | f1-score | 97.45 | 60.15 | 97.36 | 95.67 | 95.67 | 96.32 | 95.63 |
| | 200 | accuracy | 97.34 | 49.56 | 97.06 | 95.63 | 95.63 | 96.27 | 95.63 |
| | | precision | 97.21 | 67.75 | 97.30 | 96.17 | 96.17 | 96.60 | 95.63 |
| | | recall | 97.34 | 49.56 | 97.06 | 95.63 | 95.63 | 96.27 | 95.63 |
| | | f1-score | 97.05 | 54.72 | 97.08 | 95.61 | 95.61 | 96.26 | 95.63 |

**Table 5.** Experimental results of the proposed model for multiple models based on the MI feature selection method on 200, 300, 400, and 500 features

| MobileNetV2 1280 Features | Number of Selected Features | Metrics | Machine Learning Models | | | | | | |
|---|---|---|---|---|---|---|---|---|---|
| | | | SVM-rbf | SVM-poly | SVM-Linear | KNN | DT | LDA | Naïve Bayes |
| MI | 500 | accuracy | 89.84 | 87.94 | 90.87 | 82.50 | 82.50 | 92.46 | 82.50 |
| | | precision | 91.22 | 89.61 | 92.26 | 84.90 | 84.90 | 93.81 | 82.50 |
| | | recall | 89.84 | 87.94 | 90.87 | 82.50 | 82.50 | 92.46 | 82.50 |
| | | f1-score | 89.71 | 87.97 | 90.94 | 81.99 | 81.99 | 92.63 | 82.50 |
| | 400 | accuracy | 89.64 | 87.66 | 90.40 | 81.94 | 81.94 | 92.54 | 81.94 |
| | | precision | 91.06 | 89.28 | 91.69 | 84.37 | 84.37 | 93.66 | 81.94 |
| | | recall | 89.64 | 87.66 | 90.40 | 81.94 | 81.94 | 92.54 | 81.94 |
| | | f1-score | 89.47 | 87.67 | 90.43 | 81.23 | 81.23 | 92.61 | 81.94 |
| | 300 | accuracy | 88.93 | 86.55 | 90.12 | 81.07 | 81.07 | 91.87 | 81.07 |
| | | precision | 90.27 | 88.27 | 91.37 | 83.38 | 83.38 | 92.90 | 81.07 |
| | | recall | 88.93 | 86.55 | 90.12 | 81.07 | 81.07 | 91.87 | 81.07 |
| | | f1-score | 88.74 | 86.55 | 90.11 | 80.55 | 80.55 | 91.92 | 81.07 |
| | 200 | accuracy | 88.29 | 85.40 | 88.37 | 79.01 | 79.01 | 90.95 | 79.01 |
| | | precision | 89.73 | 87.26 | 89.87 | 82.15 | 82.15 | 92.23 | 79.01 |
| | | recall | 88.29 | 85.40 | 88.37 | 79.01 | 79.01 | 90.95 | 79.01 |
| | | f1-score | 88.15 | 85.47 | 88.46 | 78.62 | 78.62 | 91.04 | 79.01 |

In the last study, MobileNetV2 was used for feature extraction (Table 5). The algorithm selected for feature selection in this study is MI. As in the other studies, the number of features selected was reduced from 500 to 200. In this study, as seen in the first 2 studies, the highest accuracy rate (92.46%), recall rate (92.46%), precision rate (93.81%), and f1 score rate (92.63%) were obtained in the LDA algorithm.

In the selection of 500 features, the lowest accuracy rate of 82.50%, recall rate of 82.50%, and f1 score rate of 81.99% were obtained from KNN and DT algorithms, which are equal to each other. The lowest precision rate is 82.50%, obtained with the Naïve Bayes algorithm.

When 400 features were selected, the lowest precision, recall, and f1 score ratios were obtained with KNN and DT algorithms. The lowest precision rate was 81.94% when using the Naïve Bayes algorithm.

When 300 features were selected, the lowest accuracy and recall rates were obtained from KNN, DT, and Naïve Bayes algorithms, equal to 81.07%. The lowest precision rate was obtained with the Naïve Bayes algorithm, with a rate of 81.07%. On the other hand, the lowest f1 score ratio was obtained at 80.55% when using KNN and DT algorithms, which are equal to each other. Finally, when the feature selection was reduced to 200, it was observed that the lowest accuracy, recall, precision, and f1 score ratio were obtained with Naïve Bayes with a rate of 79.01% in general.

Figure 4 displays a comparison of the highest accuracy of the suggested model. The greatest accuracy for the LDA model using the Chi 2 feature selection approach is 92.50%, as shown in Figure 4 (a). The highest accuracy for SVM-rbf and SVM-Linear using the NCA feature selection technique is 97.70% and 97.34%, respectively, as shown in Figure 4 (b). The greatest accuracy for the LDA model when employing the MI feature selection approach is 92.54%, as shown in Figure 4 (c).

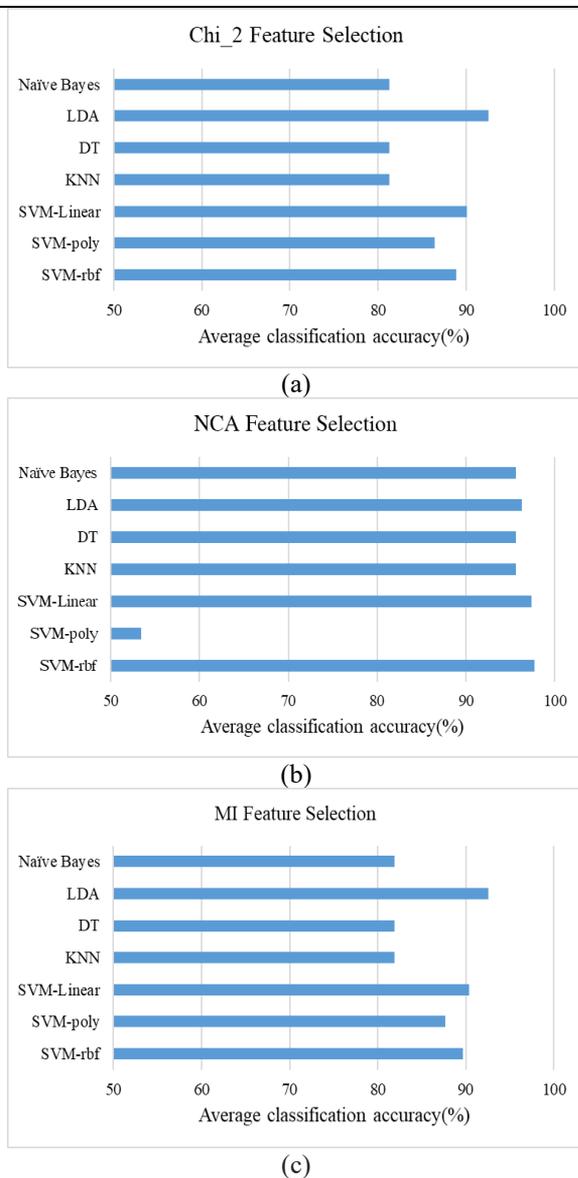

**Figure 4.** Comparison of the best model performance and accuracy using different feature selection methods ((a) Chi_2 feature selection method; (b) NCA feature selection method; (c) MI feature selection method)

NCA in the feature selection process. The lowest f1-score ratio was obtained by selecting 200 features using the Chi_2 feature selection algorithm. The ratio obtained was 75.12% in the Naïve Bayes algorithm. In general, studies show a decrease in the ratios as the number of features decreases, while in the study using NCA, it is observed that the ratios increase as the number of features increases.

## 4. CONCLUSION

The primary objective is to use transfer learning and feature selection techniques to authenticate offline handwritten signatures using the minimum acceptable features while working with a dataset that includes a substantial number of signers. Initially, a dataset comprising 420 singers is gathered, and a transfer learning model (MobileNetV2) is used to extract characteristics. To extract the acceptable features, 12 feature vectors from three feature selection techniques, including Chi_2, NCA, and MI, with 200, 300, 400, and 500 features, are used. Seven machine learning methods are used to validate the signature from specified features, including SVM-rbf, SVM-poly, SVM-linear, KNN, DT, LDA, and Naïve Bayes. The proposed model can authenticate the signature using less than 39% of all characteristics, surpassing the performance accuracy of the 97% model.

In the future, we will use several trainable feature selection strategies to identify the most suitable features accurately. These techniques will automatically determine the optimal minimal number of features. Additionally, we will utilize a graph convolutional neural network to verify offline handwritten signatures.

As a result of all the studies, when all the tables were examined, the highest accuracy rate was obtained from LDA classification using the NCA feature selection method, with a rate of 93.10% by selecting 500 features. The lowest accuracy rate was obtained in the Naïve Bayes algorithm with a rate of 75.12% using the Chi_2 feature selection algorithm, where 200 features were selected. The highest precision rate was obtained in the LDA algorithm, with 94.18%. This study was conducted with NCA by selecting 500 features. The lowest precision rate of 75.12% was obtained in Naïve Bayes with 200 feature selections using the Chi_2 feature selection algorithm. Unlike the others, the highest recall rate was observed with 200 features, which is the lowest number of feature selections made with NCA. 97.06% was obtained with the SVM-Linear algorithm. The lowest recall rate was 75.12% for 200 features. This study was conducted using the Chi-squared feature selection method. The highest f1-score ratio is 96.26%. This rate was obtained by selecting 200 features using